\documentclass{article}
\usepackage{spconf,amsmath,graphicx}
\usepackage{hyperref}
\usepackage{graphicx}

\usepackage{enumitem}
\setlist{nosep, leftmargin=14pt}

\usepackage{mwe} 
\usepackage{url} 
\usepackage{xcolor}

\title{ACS-SegNet: An Attention-Based CNN-SegFormer Segmentation Network for Tissue Segmentation in Histopathology
}
\name{
\begin{tabular}{c}
Nima Torbati$^{\star}$\thanks{Nima Torbati and Anastasia Meshcheryakova contributed equally.\\ Diana Mechtcheriakova and Amirreza Mahbod contributed equally.} \qquad
Anastasia Meshcheryakova$^{\dagger \ddagger}$ \qquad
Ramona Woitek$^{\star}$ \\
Diana Mechtcheriakova$^{\dagger \ddagger}$ \qquad
Amirreza Mahbod$^{\star}$
\end{tabular}
}

\address{$^{\star}$ Research Center for Medical Image Analysis and Artificial Intelligence,\\ Department of
Medicine, Faculty of Medicine and Dentistry, Danube Private University \\
$^{\dagger}$ Department of Pathophysiology and Allergy Research,\\ Center of Pathophysiology, Infectiology and Immunology, Medical University of Vienna \\
$^{\ddagger}$ Comprehensive Center for AI in Medicine (CAIM), Medical University of Vienna
}
\begin{document}
%
\maketitle
\begin{abstract}
Automated histopathological image analysis plays a vital role in computer-aided diagnosis of various diseases. Among developed algorithms, deep learning-based approaches have demonstrated excellent performance in multiple tasks, including semantic tissue segmentation in histological images. In this study, we propose a novel approach based on attention-driven feature fusion of convolutional neural networks (CNNs) and vision transformers (ViTs) within a unified dual-encoder model to improve semantic segmentation performance. Evaluation on two publicly available datasets showed that our model achieved \textmu IoU/\textmu Dice scores of 76.79\%/86.87\% on the GCPS dataset and 64.93\%/76.60\% on the PUMA dataset, outperforming state-of-the-art and baseline benchmarks. The implementation of our method is publicly available in a GitHub repository: \url{https://github.com/NimaTorbati/ACS-SegNet}

\end{abstract}


\section{Introduction}

\label{sec:intro}
Digital histopathology plays a crucial role in computer-aided cancer diagnosis by providing valuable information about the tumor features including the tumor microenvironment~\cite{campanella2019clinical}. In histopathological image analysis, tissue segmentation is one of the fundamental steps supporting various downstream analysis tasks. While this can be performed manually, the gigapixel resolution of whole-slide images and the need for expert knowledge make the process challenging and time-consuming. Therefore, automated computer-aided segmentation methods represent a promising alternative. In particular, deep learning–based automatic segmentation methods have emerged as powerful tools for tissue segmentation in digital pathology~\cite{atabansi2023survey}. State-of-the-art approaches are primarily built upon convolutional neural networks (CNNs)~\cite{mahbod2022dual}, vision transformers (ViTs)~\cite{horst2024cellvit}, or hybrid architectures that combine the strengths of CNNs and ViTs~\cite{tran2022trans2unet}. CNN-based models are effective in capturing local spatial features, while ViTs excel at capturing long-range contextual relationships. Hybrid models combine both architectures to deliver improved segmentation performance.

To combine the local feature extraction ability of CNNs with the global modeling capacity of ViTs, a few hybrid architectures have been proposed that integrate components from both paradigms in computational pathology. In these models, the integration of CNN and ViT modules typically occurs either in the encoder or through skip connections~\cite{tran2022trans2unet,chen2021transunet, li2024dectnet,du2025multiscale}. 
A representative example is TransUNet~\cite{chen2021transunet}, which employs a serial CNN–ViT encoder where CNN layers are followed by ViT blocks. Another widely adopted strategy is the use of parallel CNN–ViT encoders~\cite{zhang2024dual, li2024dectnet, he2024detisseg}, where two encoders process the input in parallel and are subsequently merged using various fusion schemes. Despite the progress made by the hybrid approaches, designing architectures that can fully exploit the complementary strengths of CNNs and ViTs remains an open challenge, particularly for histopathology tissue segmentation where both local features and long-range dependencies are essential.

Inspired by~\cite{torbati2025multi}, where improved tissue segmentation performance was demonstrated through ensembling two separate models, Res-UNet~\cite{zhang2018road} and SegFormer~\cite{xie2021segformer}, we propose an Attention-based CNN–SegFormer Segmentation Network (ACS-SegNet). Unlike ensemble approaches, ACS-SegNet adopts a dual-encoder design that integrates a ResNet encoder (CNN family) with a SegFormer encoder (ViT family) into a unified architecture. To maximize feature representation, the outputs of the two encoders are fused at multiple levels using the Convolutional Block Attention Module (CBAM)~\cite{woo2018cbam}, enabling the network to capture the most informative global and local features. Applied to two publicly available datasets, GCPS~\cite{ZHANG2026108398} and PUMA~\cite{10.1093/gigascience/giaf011}, we demonstrate improved semantic segmentation performance of the proposed model compared to state-of-the-art algorithms.



\begin{figure*}[h]
	\centering
	\includegraphics[width=0.95\textwidth]{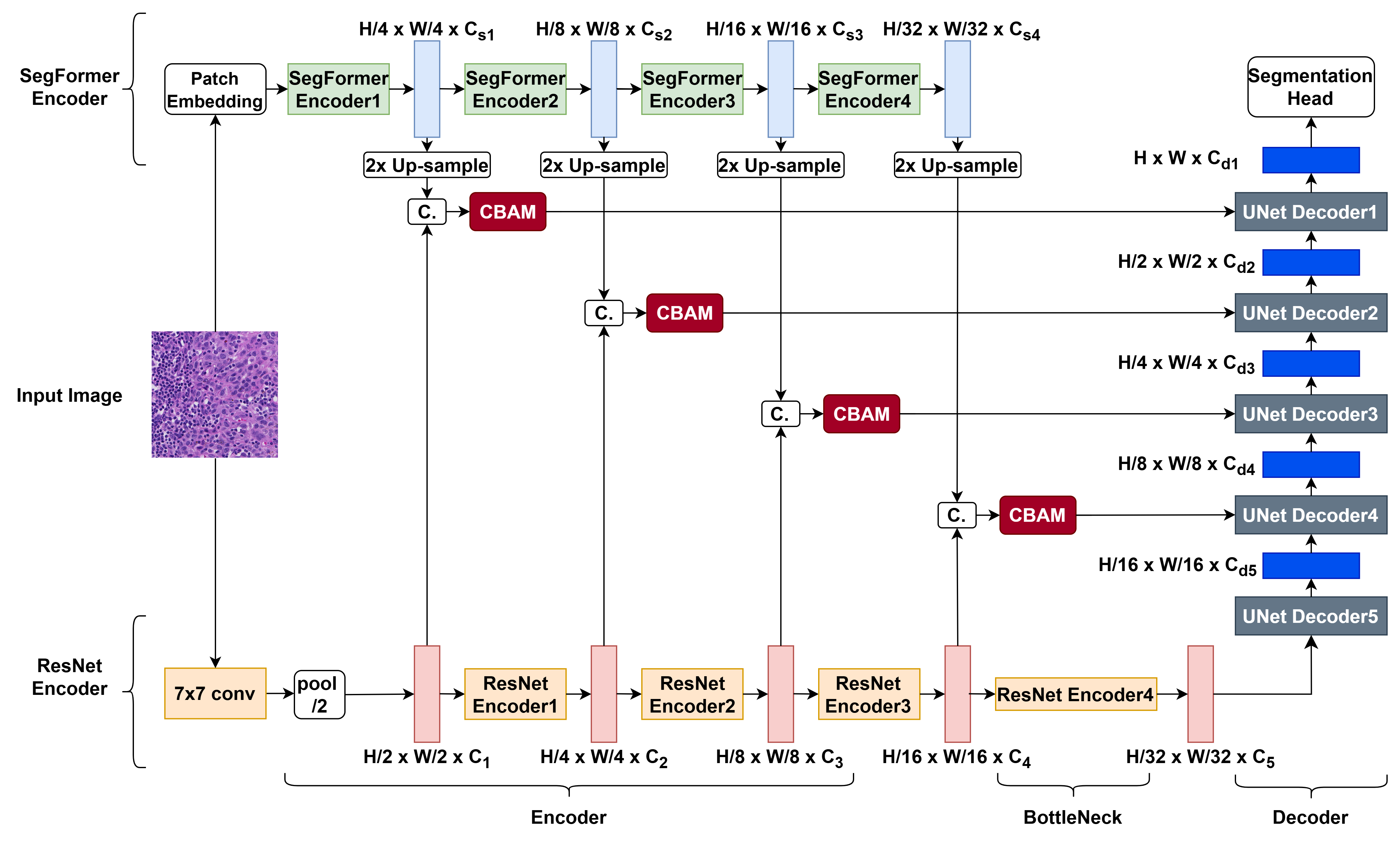}
	\caption{Block diagram of the proposed model. The model consists of two parallel encoders: a SegFormer encoder and a ResNet encoder. Features from the four stages of the SegFormer encoder are first upsampled and concatenated (C. blocks) with the corresponding ResNet encoder features, and then fused through the Convolutional Block Attention Module (CBAM). The fourth block of the ResNet encoder serves as the bottleneck. 
    For each path, the feature maps are illustrated with their corresponding dimensions. Input Image: (digital image of) H\&E-stained tissue specimen of melanoma cancer.
    }\label{fig:model}
\end{figure*}

\section{Method}
\label{sec:typestyle}

\subsection{Model}
\label{ssec:subhead}
A block diagram of the proposed model is shown in Figure~\ref{fig:model}. The model consists of two parallel encoders: a SegFormer encoder and a ResNet encoder. Features from the four stages of the SegFormer encoder are first upsampled and concatenated (C. blocks) with the corresponding ResNet encoder features, and then fused through the CBAM modules. The fourth block of the ResNet encoder serves as the bottleneck. The decoder is a traditional UNet decoder with five stages and four skip connections. The SegFormer encoder is responsible for capturing long-range dependencies, while the ResNet encoder focuses on local features. The CBAM module evaluates the importance of the feature maps and passes the weighted representations to the decoder at various levels. In our set up, the SegFormer encoder, ResNet encoder, UNet decoder, and CBAM modules are standard implementations from the respective studies~\cite{xie2021segformer, chen2018voxresnet, ronneberger2015, woo2018cbam}.



\subsection{Datasets}
\label{ssec:Datasets}
To train and evaluate the performance of the proposed method, we used the GCPS and PUMA datasets.

\subsubsection{GCPS}

The GCPS dataset contains 6,287 gastric cancer image patches, each captured at 100$\times$ magnification and sized 512$\times$512 pixels. This dataset includes annotations for two tissue types (binary segmentation): cancerous and non-cancerous. For a fair comparison and similar to the original paper~\cite{ZHANG2026108398}, we resized all images to 256$\times$256 pixels in our experiments. Further details about the GCPS dataset are provided in~\cite{ZHANG2026108398}.

\subsubsection{PUMA}
The PUMA dataset contains 206 melanoma training image patches each with a size of 1024$\times$1024 pixels. The dataset includes annotations for six tissue types (multi-class segmentation): tumor, stroma, epidermis, necrosis, blood vessel, and background. In our experiments, all training images were used except those associated with the necrosis class, as this class exhibited a strong imbalance, resulting in a total of 197 images. We excluded the official validation and test sets of the PUMA dataset, as the annotations for those images are not publicly available. To accommodate the large model size of DGAUNet~\cite{ZHANG2026108398}, a state-of-the-art segmentation model used for comparison, all images were resized to 512$\times$512 pixels due to GPU memory limitations (NVIDIA RTX 4090 GPU). Further details about the PUMA dataset are provided in~\cite{10.1093/gigascience/giaf011}.


\subsection{Experiments}


For training, we used standard shape-based transformations (flip, rotation, scaling) and intensity-based augmentations (hue, saturation, brightness, contrast) to artificially increase the dataset size.

We compared the results of our proposed approach with two state-of-the-art architectures, TransUNet~\cite{chen2021transunet} and DGAUNet~\cite{ZHANG2026108398}, as well as the baseline SegFormer~\cite{xie2021segformer} and ResNet-UNet~\cite{chen2018voxresnet} models. TransUNet employs a hybrid CNN-ViT serial encoder and DGAUNet uses two parallel CNN-based encoders. We also performed an ablation study by replacing the CBAM modules with simple concatenation (CS-SegNet) to investigate the effectiveness of the attention-based CNN–ViT feature fusion in our proposed approach. All models, except for DGAUNet, were initialized with ImageNet-pretrained weights. For SegFormer and ResNet-UNet (both as baseline models and within our proposed approach), we utilized the SegFormer-B2 and ResNet34 encoder variants, while for TransUNet, the ResNet50-B16 configuration was used.



In our experiments, we split the datasets into three folds and reported the mean and standard deviation for each model. To compare the performance of the studied methods, we employed two metrics, including the micro Intersection over Union (\textmu IoU) and micro Dice Coefficient (\textmu Dice). 
\begin{equation}
\mu IoU = \frac{\mu TP}{\mu TP + \mu FP + \mu FN}
\end{equation}
\begin{equation}
\mu Dice = \frac{2\mu TP}{2\mu TP + \mu FP + \mu FN}
\end{equation}
where TP, FP, and FN represent true positive, false positive, and false negative, respectively. For the micro metric, instead of computing the metric for each validation image and then averaging the results across all images, the entire validation dataset is 
treated as a single large image as proposed in~\cite{10.1093/gigascience/giaf011}. In this approach, the TP, FP, and FN values are computed 
cumulatively as follows:
\begin{equation}
\mu f = \sum_{n=1}^{N} f_n , \quad \text{for } f \in \{ \text{TP}, \text{FP}, \text{FN} \}
\end{equation}
where $N$ is the number of images in the validation set. We first compute the micro metric for each class, 
and then calculate the average across all classes.


\section{Results \& Discussion}

\subsection{GCPS Results}

\begin{table}[t]
\centering
\caption{Segmentation results on the GCPS dataset.}
\label{tab:gcps_results}
\begin{tabular}{lcc}
\hline
\textbf{Method} & \textbf{\textmu IoU (\%)} & \textbf{\textmu Dice (\%)} \\
\hline
DGAUNet~\cite{ZHANG2026108398}           & \(75.95 \pm 0.20\) & \(86.33 \pm 0.13\) \\
SegFormer~\cite{xie2021segformer}        & \(70.90 \pm 0.38\) & \(82.97 \pm 0.26\) \\
ResNetUNet~\cite{chen2018voxresnet}      & \(75.65 \pm 0.08\) & \(86.13 \pm 0.05\) \\
TransUNet~\cite{chen2021transunet}       & \(74.84 \pm 0.10\) & \(85.61 \pm 0.09\) \\ \hline
CS-SegNet                                & \(76.68 \pm 0.15\) & \(86.80 \pm 0.06\) \\
ACS-SegNet                               & \(\textbf{76.79} \pm \textbf{0.14}\) & \(\textbf{86.87} \pm \textbf{0.09}\) \\
\hline
\end{tabular}
\end{table}

The segmentation results of the GCPS dataset are presented in Table \ref{tab:gcps_results}. As shown, the proposed method achieves the best performance with an \textmu IoU of 76.79\% and a \textmu Dice score of 86.87\%. The second-best performance is obtained by our dual encoder model without the CBAM attention module (CS-SegNet), which confirms the positive effect of incorporating CBAM in the proposed model. Among other models, DGAUNet demonstrates the best performance, whereas ResNetUNet and TransUNet achieve comparable results, and SegFormer shows the lowest performance.

\subsection{PUMA Results}


\begin{figure}[t!]
	\centering
	\begin{tabular}{@{\hskip 0pt}c@{\hskip 4pt}c@{\hskip 4pt}c@{\hskip 0pt}}

		\includegraphics[width=2.5cm]{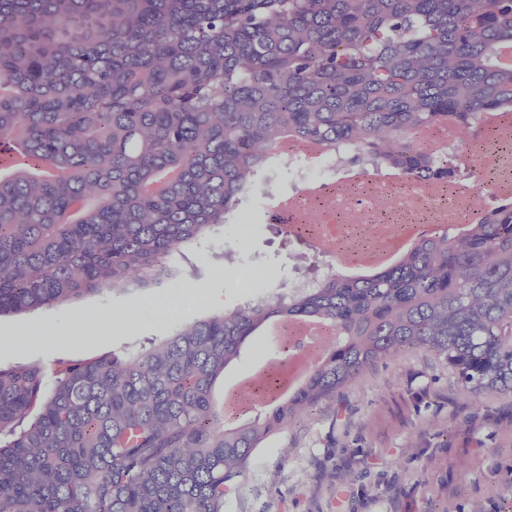} &
		\includegraphics[width=2.5cm]{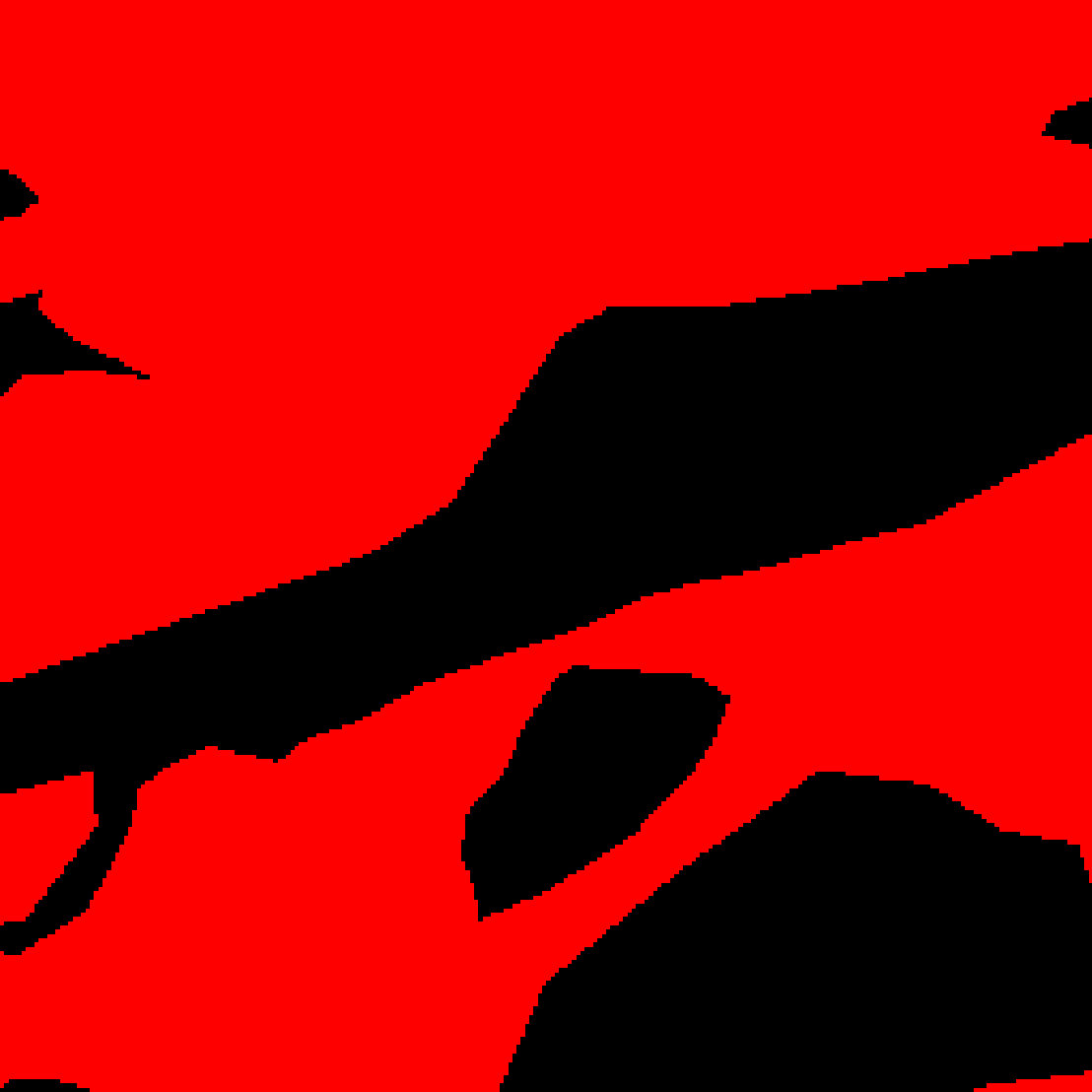} &
		\includegraphics[width=2.5cm]{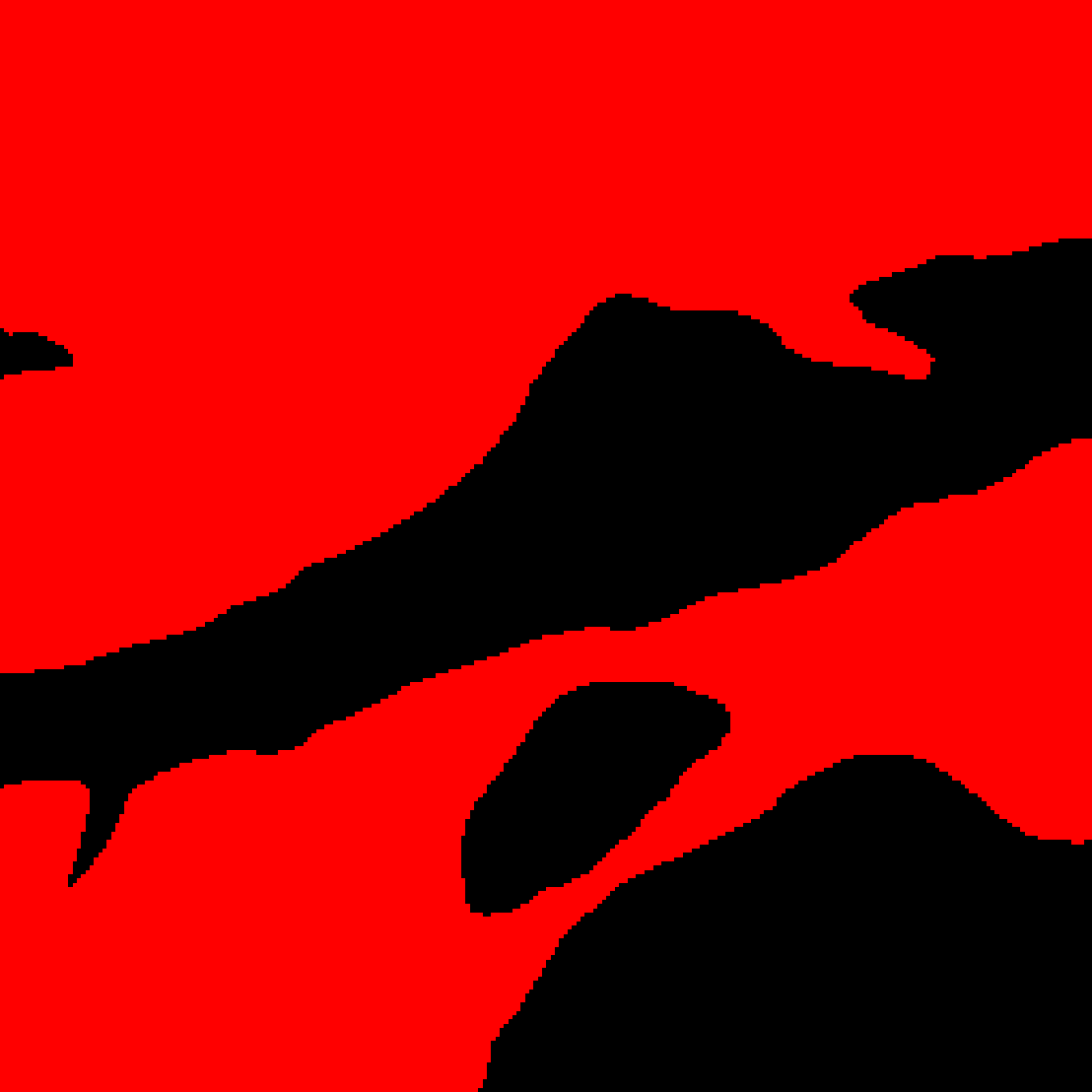} \\
		
		\includegraphics[width=2.5cm]{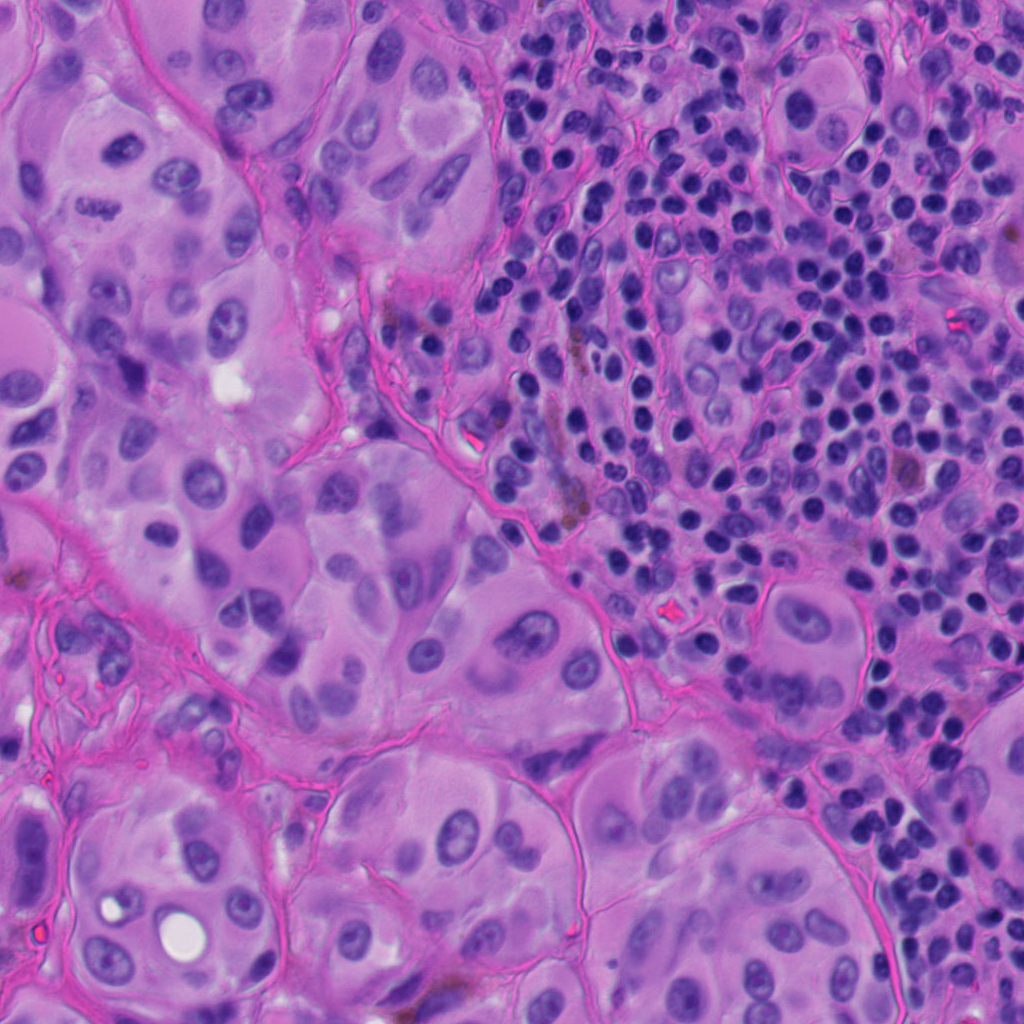} &
		\includegraphics[width=2.5cm]{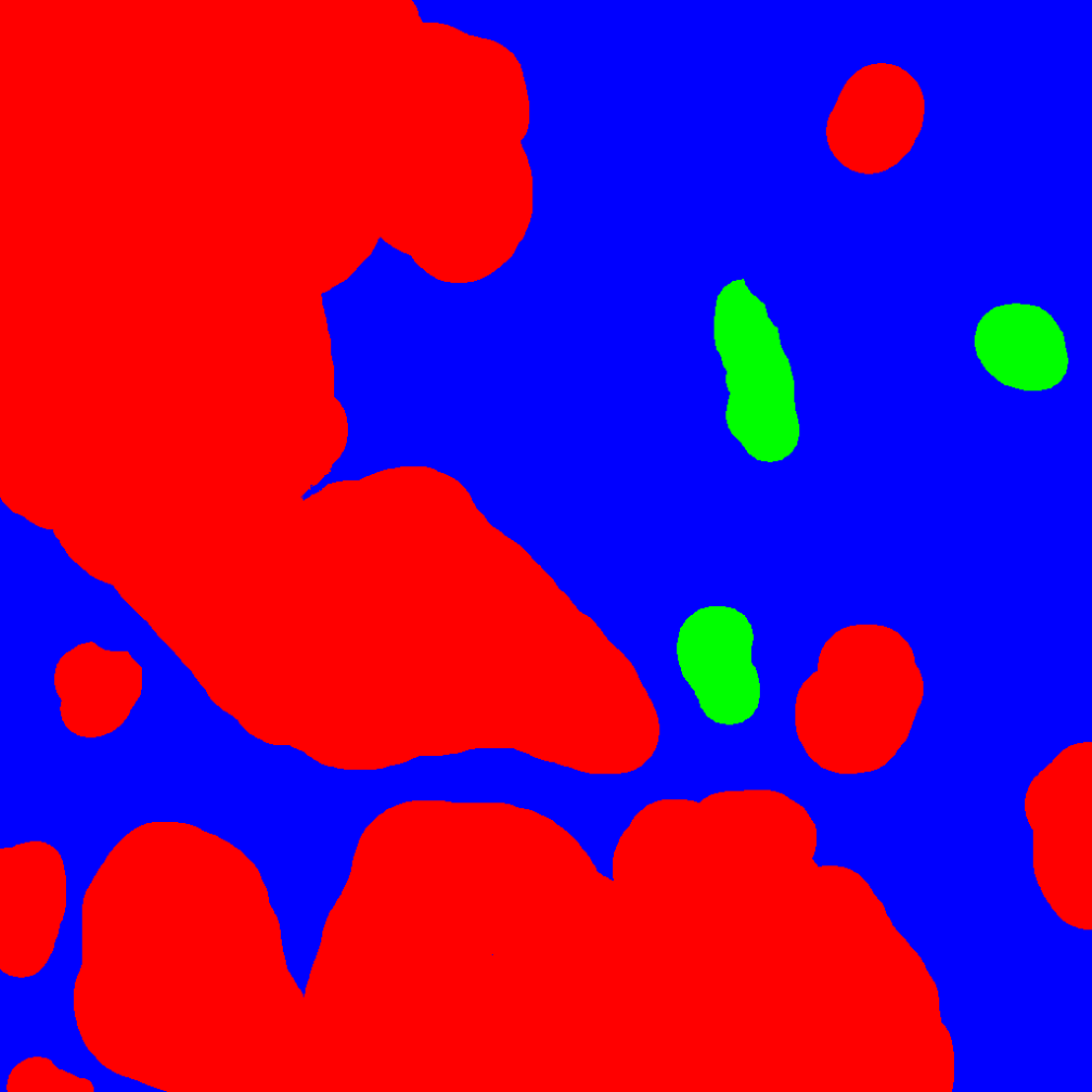} &
		\includegraphics[width=2.5cm]{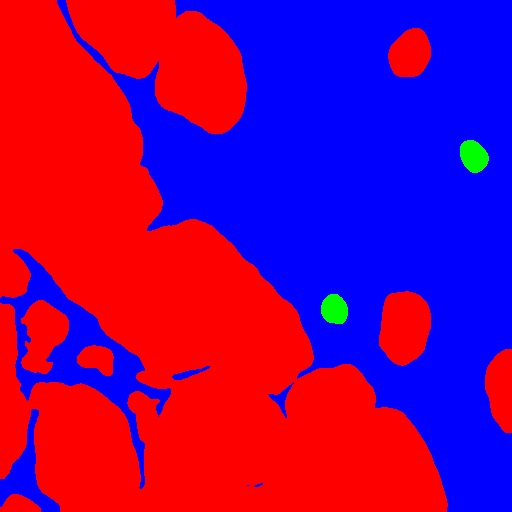}\\
        \footnotesize Input Image & \footnotesize Ground Truth & \footnotesize Prediction \\
	\end{tabular}
	\caption{Example of tissue segmentation results of the proposed method from the GCPS dataset (top) and the PUMA dataset (bottom). Tumor (cancerous) regions are shown in red, blood vessels in green, stroma in blue, and non-cancerous regions in black. Input image: (digital image of) H\&E-stained tissue specimens of gastric cancer (top) and of melanoma (bottom).}
	\label{fig:exampleT}
\end{figure}

\begin{table}[t]
\centering
\caption{Segmentation results on the PUMA dataset.}
\label{tab:puma_results}
\begin{tabular}{lcc}
\hline
\textbf{Method} & \textbf{\textmu IoU (\%)} & \textbf{\textmu Dice (\%)} \\
\hline
DGAUNet~\cite{ZHANG2026108398}           & \(44.35 \pm 1.76\) & \(53.69 \pm 1.91\) \\
SegFormer~\cite{xie2021segformer}        & \(46.78 \pm 2.58\) & \(58.25 \pm 2.01\) \\
ResNetUNet~\cite{chen2018voxresnet}      & \(58.42 \pm 1.98\) & \(71.58 \pm 1.21\) \\
TransUNet~\cite{chen2021transunet}       & \(62.43 \pm 2.47\) & \(74.63 \pm 1.91\) \\ \hline
CS-SegNet                                & \(63.67 \pm 1.23\) & \(75.55 \pm 1.71\) \\
ACS-SegNet                               & \(\textbf{64.93} \pm \textbf{2.28}\) & \(\textbf{76.60} \pm \textbf{1.36}\) \\
\hline
\end{tabular}
\end{table}

Table \ref{tab:puma_results} presents the segmentation results for the PUMA dataset. The proposed model with the CBAM attention module achieves the best performance with an \textmu IoU of 64.93\% and a \textmu Dice score of 76.60\%. The second-best performance is obtained by the proposed model without the CBAM module (CS-SegNet). Similar to the GCPS experiment, the performance of ResNetUNet is close to that of TransUNet, while SegFormer performs slightly worse. 
A noticeable difference compared to the GCPS dataset is the performance drop of DGAUNet, which fails to perform well on the multi-class segmentation task of a small-scale dataset. 

\subsection{Combined Results}
It should be noted that, in general, annotating histological images is a challenging and time-consuming task, and many annotated datasets are small-scale. Therefore, models should be capable of being trained and performing well on small-scale datasets. The experimental results indicate that the proposed model provides an optimal solution for both small- and large-scale datasets. For the GCPS dataset, which is approximately 30 times larger than PUMA, CNN-based models such as DGAUNet exhibit good performance but fail to achieve comparable results on the small-scale PUMA dataset.

However, the proposed model, which fuses features from both CNN and ViT families, achieves superior results across both datasets. For the GCPS dataset, the \textmu IoU values of DGAUNet (the second-best model for GCPS) across three folds are 75.74\%, 75.87\%, and 76.24\%, while those of the proposed method are 76.89\%, 76.62\%, and 76.78\%. These results demonstrate that the proposed method consistently achieves better performance across all folds. For the PUMA dataset—a smaller dataset involving a multi-class segmentation task—the advantage of the dual-encoder architecture becomes more evident. In this case, ACS-SegNet and TransUNet, both of which incorporate ViT components, perform better than purely CNN-based models. The \textmu IoU values of TransUNet (the second-best model for PUMA) across three folds are 64.03\%, 64.34\%, and 58.93\%, while those of the proposed method are 67.92\%, 64.51\%, and 62.37\%. These results further confirm that the proposed method consistently outperforms competing models across all folds.

The number of trainable parameters for each model is summarized in Table \ref{tab:parameters}. As shown, ACS-SegNet, with approximately one-fourth the parameter count of DGAUNet and half that of TransUNet, achieves superior performance.

Additionally to emphasize is that the superior performance of the proposed model was achieved for two different cancer types such as gastric cancer and melanoma, which diverge significantly in histological patterns.

\begin{table}[htbp]
\centering
\caption{Number of trainable parameters for the studied models.}
\begin{tabular}{l c}
\hline
\textbf{Model} & \textbf{Number of Parameters ($\sim$)} \\
\hline
DGAUNet \cite{ZHANG2026108398}  & 237 million \\
SegFormer \cite{xie2021segformer}   & 27 million \\
ResNetUNet \cite{chen2018voxresnet} & 24 million \\
TransUNet \cite{chen2021transunet}  & 105 million \\ \hline
CS-SegNet & 50 million \\
ACS-SegNet & 50 million \\
\hline
\end{tabular}
\label{tab:parameters}
\end{table}

Finally, for qualitative analysis, we show two sample results from the GSCP (top) and PUMA (bottom) datasets in Figure~\ref{fig:exampleT}. As the qualitative results demonstrate, our predicted segmentations closely match the ground-truth masks in both examples.



\section{Conclusion}
This paper introduced ACS-SegNet, an attention-based dual-encoder CNN–ViT model designed for accurate tissue segmentation in histopathology images. The architecture integrates SegFormer and ResNet encoders to effectively capture both global and local features. We evaluated the model on two publicly available histopathology datasets, GCPS and PUMA, and benchmarked its performance against state-of-the-art methods. Experimental results demonstrate that ACS-SegNet consistently outperforms other approches and achieves excellent segmentation perfromnce across both datasets.


\subsection*{Compliance with ethical standards}
\label{sec:ethics}
This research study was conducted retrospectively using human subject data made available in open access by~\cite{10.1093/gigascience/giaf011, ZHANG2026108398}. Ethical approval was not required as confirmed by the license attached with the open access data.

\subsection*{Acknowledgments}
\label{sec:acknowledgments}

This work was supported by the Vienna Science and Technology Fund (WWTF) and by the State of Lower Austria [Grant ID: 10.47379/LS23006].

\bibliographystyle{IEEEbib}
\bibliography{refs}

\end{document}